\documentclass{article}

 \usepackage[final]{corl_2017} 

\usepackage{multicol}
\usepackage{amsmath,amssymb}

\usepackage[ruled,vlined]{algorithm2e}
\usepackage{algorithmic}

\usepackage{caption}
\usepackage{graphicx}
\usepackage{epstopdf}

\usepackage{sidecap,wrapfig}
\usepackage{subfigure}
\usepackage{times}
\usepackage{microtype}
\usepackage{tabularx}
\usepackage{tikz,graphicx,units}
\usepackage{subfigure}
\usepackage{bbm}

\DeclareMathOperator*{\argmin}{arg\,min}

\newcommand{\bu}{\mathbf{u}}

\newcommand{\bx}{\mathbf{x}}

\newtheorem{theorem}{Theorem}[section]

\usepackage{url}

\usepackage{amsmath,amssymb}

\newtheorem{lemma}[theorem]{Lemma}
\newtheorem{proposition}{Proposition}[section]

\newboolean{include-notes}
\setboolean{include-notes}{True}
\newcommand{\adnote}[1]{\ifthenelse{ \boolean{include-notes}}%
 {\textcolor{red}{\textbf{#1}}}{}}
 
 \newcommand{\sknote}[1]{\ifthenelse{ \boolean{include-notes}}%
 {\textcolor{blue}{\textbf{SK: #1}}}{}}
 
  \newcommand{\mlnote}[1]{\ifthenelse{ \boolean{include-notes}}%
 {\textcolor{purple}{\textbf{ML: #1}}}{}}
 
 \newcommand{\jmnote}[1]{\ifthenelse{ \boolean{include-notes}}%
 {\textcolor{orange}{\textbf{JM: #1}}}{}}

 \newcommand{\rfnote}[1]{\ifthenelse{ \boolean{include-notes}}%
 {\textcolor{violet}{\textbf{RF: #1}}}{}}

\title{DART: \\
 Noise Injection for Robust Imitation Learning}

\author{
  Michael Laskey \\
  EECS Department\\
  UC Berkeley\\
\AND
Jonathan Lee\\
  EECS Department\\
  UC Berkeley\\
\And
Roy Fox \\
  EECS Department\\
  UC Berkeley\\
 \And
	Anca Dragan\\
  EECS Department\\
  UC Berkeley\\
\And 
	Ken Goldberg\\
  IEOR and EECS Department\\
  UC Berkeley\\
}

\begin{document}
\maketitle


\begin{abstract}
One approach to Imitation Learning is  Behavior Cloning, in which a robot observes a supervisor and infers a control policy. A known problem with this ``off-policy" approach is that the robot's errors compound when drifting away from the supervisor's demonstrations.  On-policy, techniques alleviate this by iteratively collecting corrective actions for the current robot policy. However, these techniques can be tedious for human supervisors, add significant computation burden, and may visit dangerous states during training.  We propose an off-policy approach that \emph{injects noise} into the supervisor's policy while demonstrating. This forces the supervisor to demonstrate how to recover from errors. We propose a new algorithm, DART (Disturbances for Augmenting Robot Trajectories), that collects demonstrations with injected noise, and optimizes the noise level to approximate the error of the robot's trained policy during data collection.  We compare DART with DAgger and Behavior Cloning in two domains: in simulation with an algorithmic supervisor on the MuJoCo  tasks (Walker, Humanoid, Hopper, Half-Cheetah) and in physical experiments with human supervisors training a Toyota HSR robot to perform grasping in clutter.  For high dimensional tasks like Humanoid, DART can be up to $3x$ faster in computation time and only decreases the supervisor's cumulative reward by $5\%$ during training, whereas DAgger executes policies that have $80\%$ less cumulative reward than the supervisor.  On the grasping in clutter task, DART obtains on average a $62\%$ performance increase over Behavior Cloning. 
\end{abstract}



\section{Introduction}

Robotic manipulation tasks are challenging to learn due to noncontinuous dynamics that are difficult to model, high dimensional state representations, and potentially delayed reward. Deep reinforcement learning has the potential to  train such control policies, however in practice it may require a very large number of samples~\cite{schulman2016learning}. 

Rather than pure ab initio learning, an alternative is to leverage supervision to guide the robot's policy. A common approach to this problem is Behavior Cloning, in which a robot observes a supervisor's policy and learns a mapping from state to control via regression ~\cite{pomerleau1989alvinn,bojarski2016end}. This is an \emph{off}-policy method, that suffers from compounding error when the robot executes its policy, leading it to drift to new and possibly dangerous states~\cite{ross2010reduction}.
Theoretically, the drifting occurs because of covariate shift, where execution of the robot's policy causes it to move to a different distribution from the one on which it was trained.

Ross and Bagnell proposed DAgger \cite{ross2010efficient,ross2010reduction} to correct for covariate shift by sampling states from the robot's policy. DAgger is an \emph{on}-policy method, in which the robot iteratively rolls out its current policy and asks for supervisor labels (or corrections) at the states it visits. Empirically, DAgger has been shown to reduce covariate shift and lead to robust control policies~\cite{NIPS2014_5421}. However, DAgger suffers from three key limitations: 1) it can be tedious for human supervisors to provide these corrections~\cite{laskey2016comparing} 2) it can be potentially dangerous for a physical robot to visit highly sub-optimal states ~\cite{zhang2016query}, and 3) repeatedly updating the robot's policy is computationally expensive. 
In this paper, we introduce an alternative approach to address covariate shift and enable robust policies.

 One way to expose the robot to corrective examples and reduce drift is to inject noise into the supervisor's demonstrations and let the supervisor provide corrections as needed.  Our insight is that by injecting small levels of noise, we will focus on the states that the robot needs to recover from -- the states \emph{at the boundary} of the ideal policy. This has the potential to obtain the advantages of on-policy methods by recovering from errors as the robot starts making them, without the disadvantage of aggregating these errors at training time, and getting the robot to dangerous states with low reward. We propose to still perform off-policy learning, but to inject an optimized level of noise into the supervisor's control stream as we acquire demonstrations.

Injecting noise requires selecting the magnitude and direction of the noise appropriately. Intuitively, the noise should approximate the error of the trained robot's policy, so that the demonstrations have state distributions that are close to the one that the robot will encounter at test time. This may be challenging in robotic control tasks where the control signals are high-dimensional. We thus propose to approximate the optimal noise distribution by first selecting a parameterized noise model, and then iteratively optimizing the likelihood of the supervisor's noise-injected policy applying the robot's control. We propose DART( Disturbances for Augmenting Robot Trajectories): a noise injection variant of Behavior Cloning. 

We evaluate DART on the MuJoCo locomotion environments and a grasping in clutter task on a Toyota HSR robot. In the locomotion environments, where a continuous control robot is trained to walk forward, DART achieves parity with state of the art on-policy methods, such as DAgger.  For high dimensional tasks like Humanoid, DART is $3x$ faster in computation time and during training only decreases the supervisor's cumulative reward by $5\%$, whereas DAgger executes policies that have  $80\%$ less cumulative reward than the supervisor, thus suggesting only a small amount of noise is sufficient to reduce covariate shift.  We then evaluate DART with four human supervisors, who trained a robot to perform a grasping in clutter task. In the task, a robot must reason from an eye-in-hand camera perspective how to push occluding objects away to reach a goal object. We observe that on average DART leads to a $62\%$ performance increase over traditional Behavior Cloning.

This paper contributes: 1) a new algorithm, DART, that sets the appropriate level of noise to inject into the supervisor's control stream 2) a theoretical analysis of noise injection, which illustrates when DART can improve performance over Behavior Cloning  3) experimentation with algorithmic and human supervisors characterizing noise injection can reduce covariate shift.  

\section{Related Work}

\subsection{Imitation Learning}
Imitation Learning schemes are either off-policy or on-policy. In off-policy Imitation Learning, the robot passively observes the supervisor, and  learns a policy mapping states to controls by approximating the supervisor's policy. This technique has been successful, for instance, in learning visuomotor control policies for self-driving cars~\cite{pomerleau1989alvinn,bojarski2016end}. However, Pomerleau et al. observed that the self-driving car would steer towards the edge of the road during execution and not be able to recover~\cite{pomerleau1989alvinn}. Later, Ross and Bangell theoretically showed that this was due to the robot's distribution being different than the supervisor's, a property known as covariate shift, which caused errors to compound during execution~\cite{ross2010efficient}. 

Ross et al.~\cite{ross2010reduction} proposed DAgger, an on-policy method in which the supervisor iteratively provides corrective feedback on the robot's behavior. This alleviates the problem of compounding errors, since the robot is trained to identify and fix small errors after they occur.
But despite their ability to address covariate shift, on-policy methods have been shown to be challenging for human supervisors~\cite{laskey2016comparing} and require the robot to visit potentially dangerous regions of the state space during training~\cite{zhang2016query}. Further, on-policy methods require retraining the policy from scratch after each round of corrections. 

Techniques have been proposed to address some of these limitations. Recently, Sun et al. proposed a gradient update for on-policy methods to alleviate the computation burden ~\cite{sun2017deeply}. However, it has been shown local gradient updates suffer in performance compared to full retraining of the policy~\cite{vlachos2013investigation}. Zhang et al. trained a classifier to predict when the robot is likely to error during execution and then proposed to transfer control over to the supervisor~\cite{zhang2016query}. However, this approach inherits the other limitations of on-policy methods, the challenge for human supervisors and the computational burden.

\subsection{Noise Injection to Increase Robustness}
In robust control theory, Andereson et al. noted a phenomenon similar to the covariate shift when performing model identification~\cite{sethares1989persistency}. They observed that once the model has been estimated and a control law has been optimized, the robot  visited regions of the state space in which it was unstable. A condition was proposed to correct for this known as persistance excitation, which requires the training data to be informative enough to learn a robust model ~\cite{sastry2011adaptive}. 

 One way to achieve persistance excitation is to inject isotropic Gaussian noise into the control signal~\cite{green1986persistence}. Due to its full-rank covariance matrix, such white noise can expose the system to any disturbance with positive probability~\cite{marmarelis1978white}. In light of this, we propose injecting noise into the supervisor's policy in order to collect demonstrations that simulate the errors that the robot would make at test time.


\section{Problem Statement}
The objective of Imitation Learning is to learn a policy that matches  the supervisor's policy.

\noindent\textbf{Modeling Choices and Assumptions:}  We model the system dynamics as Markovian and stochastic. We model the initial state as sampled from a distribution over the state space.
We assume a known state space and set of actions. We also assume access to a robot, such that we can sample from the state sequences induced by a policy.
Lastly, we assume access to a supervisor who can provide a demonstration of the task. 

\noindent\textbf{Policies and State Densities.}
We denote by $\mathcal{X}$ the set consisting of observable  states for a robot, and by $\mathcal{U}$ the set of actions. We model dynamics as Markovian, such that the probability of visiting state $\bx_{t+1}\in
\mathcal{X}$ can be determined from the previous state $\bx_t\in\mathcal{X}$ and action $\bu_t\in
\mathcal{U}$: 
$$p(\bx_{t+1}| \bx_{0},\bu_{0}, \ldots \bx_{t}, \bu_{t} )=p(\bx_{t+1}| \bx_t, \bu_{t}).$$
We assume a probability density over initial states $p(\bx_0)$.
An environment is thus defined as a specific instance of action and state spaces, initial state distribution, and dynamics. 


Given a time horizon $T\in \mathbb{N}$, a trajectory $\xi$ is a finite sequence of $T$ pairs of states visited and corresponding
control inputs at these states, $\xi = (\bx_0,\bu_0,\bx_1,\bu_1,\ldots, \bx_T)$, where $\bx_t\in \mathcal{X}$
and $\bu_t\in \mathcal{U}$ for each $t$.

A policy is a measurable function $\pi: \mathcal{X} \to \mathcal{U}$ from states to controls. 
We consider policies $\pi_{\theta}:\mathcal{X}\to \mathcal{U}$ parameterized by some $\theta\in \Theta$. Under our assumptions, any such policy $\pi_{\theta}$ induces a probability density over the set of  trajectories of length $T$: $$p(\xi | \pi_\theta)=
p(\bx_0)\prod_{t=0}^{T-1}\pi_\theta(\bu_t|\bx_t)p(\bx_{t+1}|\bu_t,\bx_t)$$

The term $\pi_\theta(\bu_t|\bx_t)$ indicates stochasticity in the applied policy and we consider this to be a user-defined distribution in which the deterministic output of the policy is a parameter in the distribution. An example distribution is $\epsilon$-greedy, in which with probability $\epsilon$ a random control is applied instead of $\pi_{\theta}(\bx_t)$.

While we do not assume knowledge of the distributions corresponding to $p(\bx_{t+1}|\bx_t,\bu_t)$ or $p(\bx_0)$, we assume that we have a real robot or a simulator. Therefore, when `rolling out' trajectories under a policy $\pi_{\theta}$, we utilize the robot or the simulator to sample from the resulting distribution over trajectories rather than estimating $p(\xi|\pi_\theta)$ itself.

\noindent\textbf{Objective.} In Imitation Learning, we do not assume access to a reward function, like we do in Reinforcement Learning~\cite{sutton1998reinforcement}, but instead
a supervisor, $\pi_{\theta^*}$, where $\theta^*$ may not be contained in $\Theta$. We assume the supervisor achieves a desired level of performance on the task, although it may not be optimal. 

We measure the difference between controls using a surrogate loss $l : \mathcal{U} \times \mathcal{U} \rightarrow \mathbb{R}$~\cite{ross2010reduction,ross2010efficient}.
A commonly considered surrogate loss is the squared L2-norm on the control vectors $l(\bu_1,\bu_2) = ||\bu_1- \bu_2||^2_2$~\cite{laskey2016robot}.
We measure total loss along a trajectory with respect to two policies $\pi_{\theta_1}$ and $\pi_{\theta_2}$ by $J(\theta_1, \theta_2 | \xi) = \sum_{t=0}^{T-1} l(\pi_{\theta_1}(\bx_{t}),\pi_{\theta_2}(\bx_{t}))$. 

The objective of Imitation Learning is to minimize the expected surrogate loss along the distribution induced by the robot's policy:

\begin{equation}\label{eq:main_obj}
\underset{\theta}{\mbox{min }} E_{p(\xi|\pi_\theta)} J(\theta,\theta^* | \xi).
\end{equation}

In Eq. \ref{eq:main_obj}, the distribution on trajectories and the cumulative surrogate loss are coupled, which makes this a challenging optimization problem. The field of Imitation Learning has considered two types of algorithmic solutions to this objective, off-policy learning and on-policy learning~\cite{ross2010efficient}. A common off-policy technique is Behavior Cloning, which samples from the supervisor's distribution and performs expected risk minimization on the demonstrations:

$$ \theta^R = \underset{\theta}{\mbox{argmin }} E_{p(\xi|\pi_{\theta^*})} J(\theta,\theta^* | \xi). $$

 \vspace*{-5pt}
The performance of the policy $\pi_{\theta^R}$ can be written in terms of the following decomposition:
 \vspace*{-5pt}
\begin{align*}
&E_{p(\xi |\pi_{\theta^R})} J(\theta^R,\theta^*|\xi) \\
& = \underbrace{E_{p(\xi |\pi_{\theta^R})} J(\theta^R,\theta^*| \xi) -  E_{p(\xi |\pi_{\theta^*})} J(\theta^R,\theta^*|\xi)}_{\text{Shift}} + \underbrace{E_{p(\xi |\pi_{\theta^*})} J(\theta^R,\theta^*| \xi) }_{\text{Loss}},
\end{align*}
 \vspace*{-10pt}

which corresponds to the covariate shift and the standard loss. In this work, we focus on minimizing the covariate shift. For references on how to minimize the standard loss see ~\cite{scholkopf2002learning}. In the next section, we will show how noise injection can be used to reduce covariate shift.

\section{Off-Policy Imitation Learning with Noise Injection}

\begin{wrapfigure}{r}{0.5\textwidth} \label{fig:dart_int}
    \centering
    \includegraphics[width=0.5\textwidth]{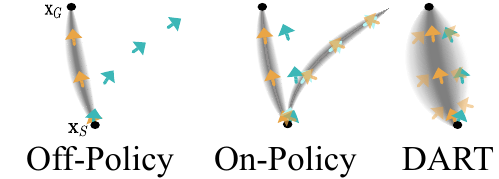}
    \caption{
    \footnotesize
    Robot Learning to reach a goal state $\bx_G$. The grey denotes the distribution over trajectories.  Left: Off-Policy learning in which the supervisor, the orange arrows, provides demonstrations. The robot, the teal arrows, deviates from the distributions and incurs high error. Middle: On-Policy which samples from the current robot's policy, the light teal arrows, to receive corrective examples from the supervisor.  Right: DART, which injects noise to widen the supervisor's distribution and provides corrective examples. DART is off-policy but robust. }
    \vspace*{-20pt}
\end{wrapfigure}

Noise injected into the supervisor's policy simulates error occurring during testing. Under this noise, the supervisor is forced to take corrective actions in order to successfully perform the task. The corrective examples allow the robot to learn how to recover when it deviates from the supervisor's distribution. However, because the demonstrations are still concentrated around the supervisor's policy, it can have less cognitive load on the supervisor than on-policy methods, which are concentrated around the robot's policy.  The intuition of providing corrective examples during data collection is shown in Fig. 1.

Denote by $p(\xi|\pi_{\theta^*},\psi)$ a distribution over trajectories with noise injected into the supervisor's distribution $\pi_{\theta^*}(\bu|\bx,\psi)$. The parameter $\psi$ represents the sufficient statistics that define the noise distribution. For example, if Gaussian noise is injected parameterized by $\psi$, $\pi_{\theta^*}(\bu|\bx,\psi) = \mathcal{N}(\pi_{\theta^*}(\bx), \Sigma)$.  

Similar to Behavior Cloning,  we can sample demonstrations from the noise-injected supervisor and minimize the expected loss via standard supervised learning techniques. This can be written as follows:

\begin{equation}\label{eq:bc_obj}
\theta^R = \underset{\theta}{\mbox{argmin }} E_{p(\xi|\pi_{\theta^*},\psi)} J(\theta,\theta^* | \xi)
\end{equation}

Eq. \ref{eq:bc_obj}, though, does not explicitly minimize the covariate shift for arbitrary choice of  $\psi$.  In order to reduce covariate shift, we introduce the following Lemma to bound the shift. We can then attempt to optimize this bound with respect to $\psi$. This bound holds when the surrogate loss $\forall \bu_1, \bu_2 \in \mathcal{U}$ $l(\bu_1,\bu_2) \in [0,1]$. Examples of when this occurs is if the robot has a discrete set of actions or  bounded continuous controls and they are normalized during learning.

\begin{lemma}\label{lm:ni}
If $\forall\bu_1, \bu_2 \in \mathcal{U}, \:\: 0 \leq l(\bu_1,\bu_2) \leq 1$ the following is true for a time horizon of $T$
$$|E_{p(\xi|\pi_{\theta^*},\psi)} J(\theta^R,\theta^* | \xi) - E_{p(\xi|\pi_{\theta^R})} J(\theta^R,\theta^* | \xi)| \leq T \sqrt{ \frac{1}{2}\mathcal{D_{KL}} \big(p(\xi|\pi_{\theta^R}),p(\xi|\pi_{\theta^*},\psi)\big)}$$
[See Appendix for Proof]
\end{lemma}

 \vspace*{-5pt}
To minimize the covariate shift,  we can optimize the KL-divergence, $\mathcal{D_{KL}}$, with respect to the sufficient statistics, $\psi$, of the noise distribution.  This can be reduced to the following optimization problem:

$$\underset{\psi}{\mbox{min}} \: E_{p(\xi|\pi_{\theta^R})} -\sum^{T-1}_{t=0} \: \mbox{log} [\pi_{\theta^*}(\pi_{\theta^R}(\bx_t)|\bx_t,\psi)]$$

 \vspace*{-5pt}

This above optimization problem decreases the negative log-likelihood of the  robot's control during data collection. Thus, we want to choose a noise parameter that makes the supervisor's distribution closer to the final robot's distribution. A clear limitation of this optimization problem is that it requires knowing the final robot's distribution $p(\xi|\pi_{\theta^R})$, which is determined only after the data is collected.  The dependency on this term creates a  \emph{chicken and egg} problem. In the next section, we present DART, which applies an iterative approach to the optimization.

\subsection{DART: Disturbances for Augmenting Robot Trajectories}
The above objective cannot be solved because $p(\xi|\pi_{\theta^R})$ is not known until after the robot has been trained.  We can instead iteratively sample from the supervisor's distribution with the current noise parameter, $\psi_k$, and minimize the negative log-likelihood of the noise-injected supervisor taking the current robot's, $\pi_{\hat{\theta}}$, control. 

\begin{equation}\label{eq:dart_max}
\hat{\psi}_{k+1} = \underset{\psi}{\mbox{argmin}} \: E_{p(\xi|\pi_{\theta^*}, \psi_k)} -\sum^{T-1}_{t=0}\mbox{log} \: [\pi_{\theta^*}(\pi_{\hat{\theta}}(\bx_t)|\bx_t,\psi)]
\end{equation}

 \vspace*{-5pt}

However, the above iterative process can be slow to converge because  it is optimizing the noise with respect to the current robot's policy and are optimizing on samples from the supervisor's distribution. We can obtain a better estimate by observing that the supervisor should simulate as much expected error as the final robot policy, $E_{p(\xi|\pi_{\theta^R})} J(\theta^R,\theta^*|\xi)$. It is possible that we have some knowledge of this quantity from previously training on similar domains.

Inspired by shrinkage estimation~\cite{daniels2001shrinkage}, we can obtain a better estimate of $\hat{\psi}$ by a target value based on the anticipated final robot's error and subsequently scaling the current simulated error to this level. Denote  $E_{p(\xi|\pi_{\theta^*}, \hat{\psi})} \sum^{T-1}_{t=0} l(\bu_t, \pi_{\theta^*}(\bx_t))$, the expected deviation from the supervisor's control for a given surrogate loss. The scaling can be denoted

\begin{equation}\label{eq:dart_shrink}
\psi^{\alpha}_{k+1} = \hat{\psi}_{k+1} * \underset{\beta\geq 0}{\mbox{argmin}} \: |\alpha- E_{p(\xi|\pi_{\theta^*}, \beta*\hat{\psi}_{k+1})} \sum^{T-1}_{t=0} l(\bu_t, \pi_{\theta^*}(\bx_t))|
\end{equation}
 \vspace*{-5pt}

where $\alpha$ corresponds to an estimated guess of the robot's final error and $\psi^{\alpha}_{k+1}$ corresponds to the sufficient statistic after the scaling has been applied. For Gaussian noise with covariance matrix $\Sigma$, the expected deviation has a closed form solution for the squared Euclidean loss: $E_{p(\xi|\pi_{\theta^*}, \hat{\psi})} \sum^{T-1}_{t=0} l(\bu_t, \pi_{\theta^*}(\bx_t))) = T\mbox{tr}(\Sigma)$.  Thus, a known  closed  form solution can be derived for both Eq. 3 and Eq. 4:

 $$\Sigma^\alpha_{k+1} = \frac{\alpha}{T\mbox{tr}(\hat{\Sigma}_{k+1})} \hat{\Sigma}_{k+1}, \: \: \: \:  \: \:\hat{\Sigma}_{k+1}= \frac{1}{T}E_{p(\xi|\pi_{\theta^*}, \Sigma_{k}^\alpha)}\sum^{T-1}_{t=0} ( \pi_{\hat{\theta}}(\bx_t) - \pi_{\theta^*}(\bx_t))( \pi_{\hat{\theta}}(\bx_t) - \pi_{\theta^*}(\bx_t))
^T.$$
 \vspace*{-5pt}
\begin{wrapfigure}{r}{0.5\textwidth} \label{fig:dart_alg}
   \centering
   \includegraphics[width=0.5\textwidth]{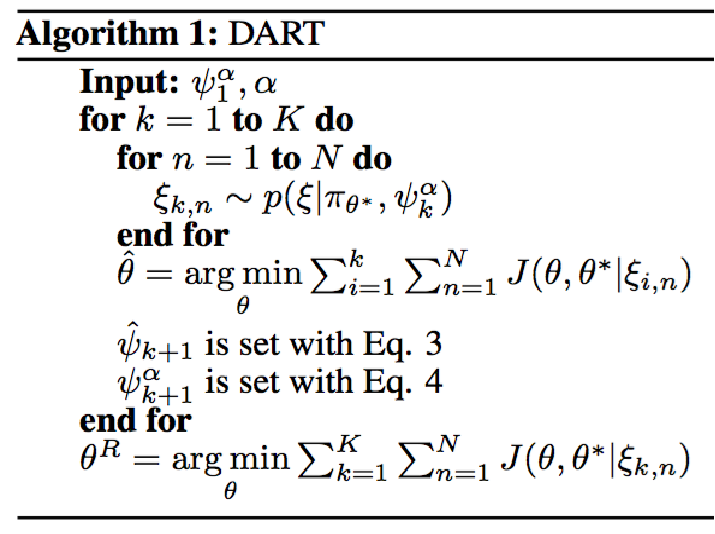}
    \vspace*{-20pt}

\end{wrapfigure}

 In Algorithm 1, we present DART, which iteratively solves  for $\psi^\alpha_{k+1}$ and to collect data and train the final robot policy.   First $N$ demonstrations are collected from a supervisor with an initial noise parameter set. Then a policy, $\pi_{\hat{\theta}}$, is learned via empirical risk minimization. The learned policy is then used to optimize Eq. \ref{eq:dart_max} based on sample estimates and the outcome is then scaled based on Eq. \ref{eq:dart_shrink}. Once the noise term is found $N$ demonstrations are collected with the noise-injected supervisor and the robot is trained on the aggregate dataset. The algorithm is repeated for $K$ iterations.

\subsection{Theoretical Analysis of DART}
Compared to Behavior Cloning, DART reduces covariate shift by simulating errors the robot is likely to make. We show the following Proposition to provide better intuition for when this improvement occurs. 

\begin{proposition}\label{thm:ni}
Given a deterministic supervisor's policy with Gaussian noise injected, $\pi^*(\bu|\bx,\psi) = \mathcal{N}(\pi^*(\bx),\Sigma)$, when the robot policy has error $E_{p(\xi|\pi_{\theta^R})} J(\theta^R,\theta^*|\xi) > 0$. The following holds:

$$ \mathcal{D_{KL}} \big(p(\xi|\pi_{\theta^R}),p(\xi|\pi_{\theta^*},\Sigma)\big) < \mathcal{D_{KL}} \big(p(\xi|\pi_{\theta^R}),p(\xi|\pi_{\theta^*})\big)$$

where the right hand side corresponds to Behavior Cloning. 
[See Appendix for Proof]
\end{proposition}

Proposition \ref{thm:ni} shows that DART reduces an upper bound on covariate shift, or the distance between the distributions, more so than Behavior Cloning, when the robot has non-zero error with respect to the supervisor. We note this assumes the supervisor is not affected by the noise injection, which for human supervisor's this might not always be true.  Thus, noise should be injected if the robot is expected to make some errors after training. 

However, noise injection will offer no improvement if the robot can represent the supervisor perfectly and collect sufficient data.  A similar result was shown in Laskey et al.~\cite{laskey2016comparing} for when DAgger is expected to improve over Behavior Cloning. In practical applications, it is unlikely to obtain sufficient data to perfectly represent the supervisor, which highlights the need for DART.


\section{Experiments}
Our experiments are designed to explore:
1) Does DART reduce covariate shift as effectively as on-policy methods?
2) How much does DART reduce the computational cost and how much does it decay the supervisor's performance during data collection?
3) Are human supervisors able to provide better demonstrations with DART?

\begin{figure}[h]
\includegraphics{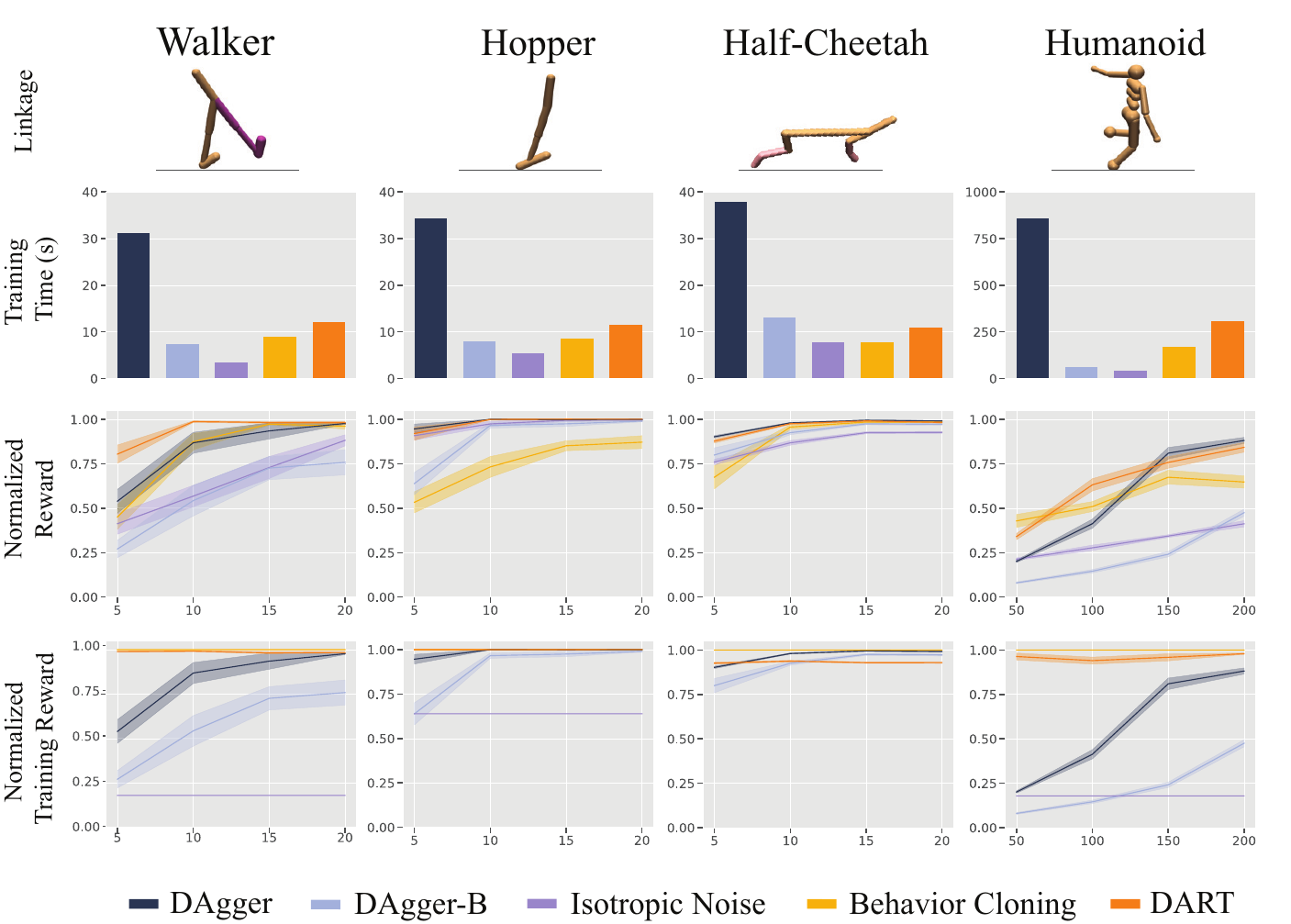}
\centering
\caption{
    \footnotesize
Top: The four different locomotive domains in MuJoCo we evaluated DART on: Walker, Hopper, Half-Cheetah and Humanoid. Top Middle: The time, in seconds, to achieve the performance level reported below. DART achieves similar performance to DAgger in all 4 domains, but requires significantly less computation because it doesn't require retraining the current robot policy after each demonstration. DAgger-B reduces the computation required by less frequently training the robot, but suffers significantly in performance in domains like Humanoid. Bottom Middle: The learning curve with respect to reward obtained during training with each algorithm plotted across the number of demonstrations. Bottom: The reward obtained during collection of demonstrations for learning. DART receives  near the supervisor's reward at all iterations whereas DAgger can be substantially worse in the beginning. }
\vspace*{-15pt}
\label{fig:mujoco}
\end{figure}

\label{sec:result}
\subsection{MuJoCo Locomotion Environments}
To test how well DART performs compared with an on-policy methods such as DAgger and off-policy methods like Behavior Cloning, we use MuJoCo locomotion environments~\cite{todorov2012mujoco}.  The challenge for these environments is that the learner does not have access to the dynamics model and must learn a control policy that operates in a high-dimensional continuous control space and moves the robot in a forward direction without falling. 

We use an algorithmic supervisor in these domains: a policy which is trained with TRPO~\cite{schulman2015trust} and is represented as a neural network with two hidden layers of size $64$. Note, while TRPO uses a stochastic policy, the supervisor is the deterministic mean of the learner. This is the same supervisor used in~~\cite{ho2016generative}. For all 4 domains, we used the same neural network as the supervisor and trained the policies in Tensorflow. At each iteration we collected one demonstration. In order to make the task challenging for Imitation Learning, we used the same technique as in~\cite{ho2016generative}, which is to sub-sample $50$ state and control pairs from the demonstrated trajectories, making the learners receive less data at each iteration.

We compare the following five techniques: Behavior Cloning, where data is collected from the supervisor's distribution without noise injected; DAgger, which is an on-policy method that stochastically samples from the robot's distribution and then fully retrains the robot's policy after every demonstration is collected; DAgger-B, which is DAgger but the retraining is done intermittently to reduce computation time; Isotropic-Gaussian which injects non-optimized isotropic noise; and DART. See supplement material for how the hyper-parameters for each method are set. 

We used the following MuJoCo environments:  Walker, Hopper, Humanoid, and Half-Cheetah. The size of the state and control space for each task is $\lbrace |\mathcal{X}| = 17, |\mathcal{U}| = 6 \rbrace$, $\lbrace |\mathcal{X}| = 11, |\mathcal{U}| = 3 \rbrace$, $\lbrace |\mathcal{X}| = 376, |\mathcal{U}| = 17 \rbrace$, $\lbrace |\mathcal{X}| = 117, |\mathcal{U}| = 6 \rbrace$, respectively. All experiments were run on a MacBook Pro with an 2.5 GHz Intel Core i7 CPU.  We measured the cumulative reward of each learned policy by rolling it for $500$ timesteps, the total computation time for each method and the cumulative reward obtained during learning.

Fig. \ref{fig:mujoco} shows the results. In all domains, DART achieves parity with DAgger, whereas Behavior Cloning and DAgger-B are below this performance level in Walker and Humanoid.  For Humanoid, DART is $3x$ faster in computation time and during training only decreases the supervisor's cumulative reward by $5\%$, whereas DAgger executes policies that have over $80\%$ less cumulative reward than the supervisor. The reason for this is that DAgger requires constantly updating the current robot policy and forces the robot into sub-optimal states.  While, one way to reduce this computation is to decrease the number of times the policy is updated.  DAgger-B illustrates that this can significantly deteriorate performance. Lastly, naively applying isotropic noise does not perform well, and leads to unsafe policies during execution, which suggests the need for optimizing the level of noise. 

 \vspace*{-10pt}

\subsection{Robotic Grasping in Clutter}

\begin{figure}
\center
\includegraphics{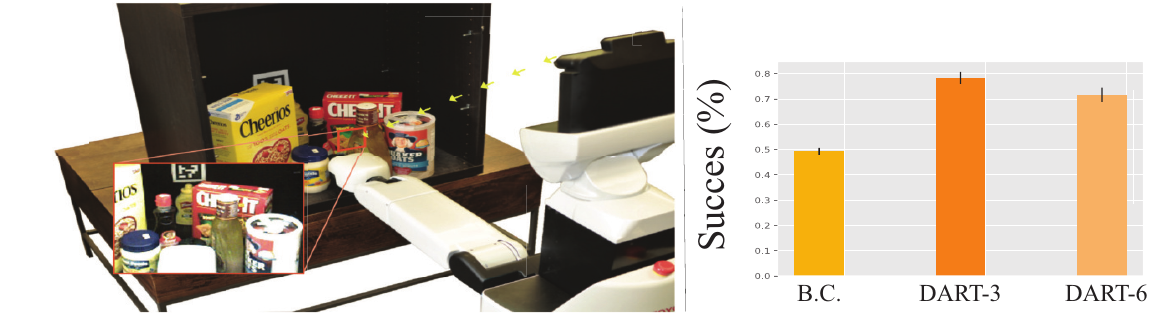}
\caption{
    \footnotesize
Left: Experimental setup for the grasping in clutter task. A Toyota HSR robot uses a head-mounted RGBD camera and  its arm to push obstacle objects out of the way to reach the goal object, a mustard bottle. The robot's policy for pushing objects away uses a CNN trained on images taken from the robot's Primesense camera, an example image from the robot's view point is shown in the orange box. Right: the Success Rate for Behavior Cloning,  DART($\alpha=3$) and DART($\alpha =6$). DART($\alpha=3$) achieves the largest success rate. }
\vspace*{-20pt}
\label{fig:human_study}
\end{figure}

We evaluate DART with human supervisors in a grasping in clutter task on a Toyota HSR robot. We consider a  task inspired by the Amazon Picking Challenge~\cite{correll2016lessons}.  The goal of the task is for the robot to retrieve a goal object from a cupboard. The task is challenging because the goal object is occluded by obstacle objects and the robot must reason about how to clear a path based on observations taken from an eye-in-hand perspective. The objects are 6 common household food items, which consist of boxes and bottles with varying textures and mass distributions. The target object is fixed to always be a mustard bottle. The robot, task and image viewpoint are shown in Fig. \ref{fig:human_study}. See supplement material for additional information on the task. 

We use 4 supervisors who have robotics experience but not specifically in the field of Imitation Learning and compare Behavior Cloning and DART.  When performing the study, we first collect $N=10$ demonstrations with Behavior Cloning (i.e. no noise) and then in a counter-balanced ordering collect $N=30$ more demonstrations with each technique.  Our experiment was within-subject, have every supervisor perform all three methods. 

We only updated the noise parameter after the first 10 demonstrations (i.e. $K=2$).  The final robot policy was then trained on the total of $40$ demonstrations. 
We consider two different choices of  $\alpha$: $\alpha = 3T\mbox{tr}(\hat{\Sigma_1})$ and $\alpha = 3T\mbox{tr}(\hat{\Sigma_1})$. These choices correspond to the intuition that with small datasets in high dimensional image space the robot will have significant shift from the current loss on the supervisor's distribution. Through the rest of the paper, we will write our choices as $\alpha = 3$ and $\alpha = 6$ for brevity in notation. 

During policy evaluation, we measured success as 1) the robot is able to identify the goal object and 2) there being a clear path between its gripper and the object. Once these conditions are identified, as detailed in the supplement material,  an open-loop motion plan is generated to execute a grasp around the target object. 

In Fig. \ref{fig:human_study}, we report  the average performance of the three techniques across DART 20 times on different sampled initial states.  DART with $\alpha = 3$ performs the best out of the three techniques, with a $79\%$ success rate over traditional Behavior Cloning's $49\%$ success rate. Interestingly, DART with $\alpha = 6$ performs better than Behavior Cloning, but only has a $72\%$ success rate. This may suggest this level of noise was potentially too high for the human supervisor. 

 \vspace*{-10pt}

\section{Conclusion}

 \vspace*{-10pt}
This paper considers injecting noise into the supervisor's demonstrations to minimize covariate shift.  Our algorithm, DART, provides the robot with corrective examples at the boundary of the supervisor's policy. By being concentrated around the supervisor's policy, it collects demonstrations without visiting highly sub-optimal states and is easier for human supervisors tele-operating. We demonstrate it achieves parity with on-policy methods in simulated MuJoCo domains and is significantly better than traditional Behavior Cloning on a real robot. For code and more information see \url{http://berkeleyautomation.github.io/DART/}

\section{Acknowledgements}

This research was performed at the AUTOLAB at UC Berkeley in
affiliation with the Berkeley AI Research (BAIR) Lab, the Berkeley
Deep Drive (BDD) Initiative, the Real-Time Intelligent Secure
Execution (RISE) Lab, and the CITRIS "People and Robots" (CPAR)
Initiative and by the Scalable Collaborative Human-Robot Learning (SCHooL)
Project, NSF National Robotics Initiative Award 1734633. Additional donations were from donations from Toyota, Siemens, Google, Cisco, Samsung, Autodesk, Intuitive Surgical, Amazon,  and IBM. Any opinions, findings, and conclusions or recommendations expressed in this material are those of the author(s) and do not necessarily reflect the views of the Sponsors. We thank our colleagues who provided helpful feedback and suggestions, in particular Kevin Jamieson, Sanjay Krishnan, Jeff Mahler and Zoey McCarthy. 

\bibliographystyle{plainnat}
\bibliography{references}

\section{Appendix}

\subsection{Proofs for Theoretical Analysis}
\setcounter{section}{4}
\setcounter{theorem}{0}

\begin{lemma}\label{lm:ni}
If $\forall\bu_1, \bu_2 \in \mathcal{U}, \:\: 0 \leq l(\bu_1,\bu_2) \leq 1$, the following is true: 
$$|E_{p(\xi|\pi_{\theta^*},\psi)} J(\theta,\theta^* | \xi) - E_{p(\xi|\pi_{\theta})} J(\theta,\theta^* | \xi)| \leq T \sqrt{ \frac{1}{2}\mathcal{D_{KL}} \big(p(\xi|\pi_{\theta}),p(\xi|\pi_{\theta^*},\psi)\big)}$$

\end{lemma}

\setcounter{theorem}{0}

\emph{Proof:}\\

\begin{align}
&|E_{p(\xi|\pi_{\theta^*},\psi)} J(\theta,\theta^* | \xi) - E_{p(\xi|\pi_{\theta})} J(\theta,\theta^* | \xi)| \\
&\leq T||p(\xi|\pi_{\theta^*},\psi) - p(\xi|\pi_{\theta})||_{TV}\\
& \leq T \sqrt{\frac{1}{2} \mathcal{D_{KL}} \big(p(\xi|\pi_{\theta}),p(\xi|\pi_{\theta^*},\psi)\big)}
\end{align}

The first line of the proof follos from Lemma 4.2, which is stated below and the fact $\forall\bu_1,\bu_2 \:\: 0 \leq l(\bu_1,\bu_2) \leq 1$. The second line follows from Pinsker's Inequality. 
$\blacksquare$

\setcounter{theorem}{1}
\begin{lemma}\label{lem:tv_dist}
Let $P$ and $Q$ be any distribution on $\mathcal{X}$. Let $f:\mathcal{X} \rightarrow [0,B]$. Then
$$|E_{P}[f(x)] - E_Q[f(x)]| \leq B ||P-Q||_{TV}$$
\end{lemma}

\setcounter{theorem}{1}

\emph{Proof:}\\

\begin{align}
|E_{P}[f(x)] - E_Q[f(x)]| & = |\int_x p(x)f(x)dx - \int_x q(x)f(x)dx| \nonumber \\
&= |\int_x (p(x)-q(x))f(x)dx|  \nonumber \\
&=|\int_x \big(p(x)-q(x)\big)(f(x) - \frac{B}{2})dx \nonumber \\
&+ \frac{B}{2}\int_x \big(p(x) - q(x)\big)dx| \nonumber \\
&\leq \int_x |p(x)-q(x)||f(x) - \frac{B}{2}|dx \nonumber \\
&\leq \frac{B}{2} \int_x |p(x)-q(x)|dx\nonumber \\
& \leq B ||P-Q||_{TV} \nonumber
\end{align}

The last line applies the definition of total variational distance, which is $||P-Q||_{TV} = \frac{1}{2} \int_x |p(x)-q(x)|$.
$\blacksquare$

\setcounter{theorem}{0}
\begin{proposition}
Given a supervisor's policy with Gaussian noise injected, $\pi^*(\bu|\bx,\psi) = \mathcal{N}(\pi^*(\bx),\Sigma)$, when the robot policy has error $E_{p(\tau|\pi_{\theta})} J(\theta,\theta^*|\xi) > 0$. The following is true:

$$ \mathcal{D_{KL}} \big(p(\xi|\pi_{\theta}),p(\xi|\pi_{\theta^*},\Sigma)\big) < \mathcal{D_{KL}} \big(p(\xi|\pi_{\theta}),p(\xi|\pi_{\theta^*})\big)$$

where the right hand side corresponds to Behavior Cloning. 
\end{proposition}

\setcounter{theorem}{0}

\emph{Proof:} We begin the proof with the definition of the KL-divergence: 
\begin{align}
 \mathcal{D_{KL}} \big(p(\xi|\pi_{\theta}),p(\xi|\pi_{\theta^*})\big)
 =& E_{p(\xi|\pi_{\theta})} \: \mbox{log} \frac{\prod^{T-1}_{t=0} \pi_{\theta}(\bu_t|\bx_t)}{\prod^{T-1}_{t=0} \pi_{\theta^*}(\bu_t|\bx_t)}\\
  =& E_{p(\xi|\pi_{\theta})} \: \sum^{T-1}_{t=} \mbox{log} \frac{\pi_{\theta}(\bu_t|\bx_t)}{ \pi_{\theta^*}(\bu_t|\bx_t)}
\end{align}

We know that if $E_{p(\tau|\pi_{\theta})} J(\theta,\theta^*|\psi) = 0$, then the KL divergence would be zero because at all states under the distribution the two policies would perfectly agree. However, if they do not agree then the KL-divergence becomes the there $ \exists \xi$ such that $p(\xi|\pi_{\theta*}) = 0$ when $p(\xi|\pi_{\theta}) > 0$, which implies~\cite{lesne2014shannon}.

$$ \mathcal{D_{KL}} \big(p(\xi|\pi_{\theta}),p(\xi|\pi_{\theta^*})\big) = \infty$$

One technique to correct for this is to ensure that the supervisor has finite probability of applying any control (i.e. $\forall \xi \:  p(\xi|\pi_{\theta*}) > 0$ when $p(\xi|\pi_{\theta}) > 0$.  Injecting Gaussian noise ensures non-zero probability for all controls at a given trajectory and subesequently: 

$$ \mathcal{D_{KL}} \big(p(\xi|\pi_{\theta}),p(\xi|\pi_{\theta^*},\Sigma)\big) < \infty$$
 
$\blacksquare$

\subsection{Derivation for Optimizing Noise}
When optimizing noise, we need to first optimize Eq. \ref{eq:dart_max}, which finds the noise parameter that maximizes the likelihood of the current robot policy and then we can scale the parameter to our prior of the error in the final robot's policy. 

In this section, we re-derive the closed from solutions for the Gaussian and $\epsilon$-greedy scenario for the MLE objective of the current robot's control. We then derive a solution for Eq. \ref{eq:dart_shrink} for the Gaussian case. 

\noindent \textbf{Gaussian MLE} We begin with the continuous case where the supervisor's policy is stochastic with distribution $\pi_{\theta^*}(\bu_t | \bx_t, \Sigma)$ defined by $\mathcal N (\pi_{\theta^*}(\bx_t, \Sigma)$ and the learner's policy is given by $\pi_{\hat \theta}$. Then the optimization problem is
\begin{align*}
    \hat \Sigma & = \argmin_{\Sigma} - E_{p(\xi | \pi_{\theta^*})} \sum_{t = 0}^{T - 1}\log \left[ \pi_{\theta^*}( \pi_{\hat \theta} (\bx_t) | \bx_t, \Sigma )  \right] \\
    & = \argmin_{\Sigma} E_{p(\xi | \pi_{\theta^*})} - \frac{T}{2}  \log \det \Sigma^{-1} + \frac{1}{2}\sum_{t = 0}^{T - 1} \left(\pi_{\hat \theta}(\bx_t) - \pi_{\theta^*}(\bx_t) \right)^T \Sigma^{-1} \left( \pi_{\hat \theta}(\bx_t) - \pi_{\theta^*}(\bx_t) \right).
\end{align*}
We take the derivative with respect to $\Sigma$ and set it equal to zero:
\begin{align*}
     - T  \Sigma + E_{p(\xi | \pi_{\theta^*})} \sum_{t = 0}^{T - 1} \left(\pi_{\hat \theta}(\bx_t) - \pi_{\theta^*}(\bx_t) \right) \left( \pi_{\hat \theta}(\bx_t) - \pi_{\theta^*}(\bx_t) \right)^T.
\end{align*}
Then the optimal covariance matrix is 
\begin{align*}
    \hat \Sigma & = \frac{1}{T} E_{p(\xi | \pi_{\theta^*})} \sum_{t = 0}^{T - 1} \left(\pi_{\hat \theta}(\bx_t) - \pi_{\theta^*}(\bx_t) \right) \left( \pi_{\hat \theta}(\bx_t) - \pi_{\theta^*}(\bx_t) \right)^T. 
\end{align*}

\noindent \textbf{ $\epsilon$-greedy MLE} For the $\epsilon$-greedy case in the discrete action domain with a finite number of controls $K$, the probability of the supervisor choosing a control at a particular state is given by
\begin{align*}
    \pi_{\theta^*}(\bu_t | \bx_t, \epsilon) & = \begin{cases}
        1 - \epsilon & \pi_{\theta^*} (\bx_t) = \bu_t \\
        \frac{\epsilon}{K - 1} & \text{otherwise}
    \end{cases}.
\end{align*}

For any $\bu_1, \bu_2 \in \mathcal U$, let the loss function $l(\bu_1, \bu_2)$ be defined as the indicator function so that $l(\bu_1, \bu_2) = 0$ for $\bu_1 = \bu_2$ and 1 otherwise. Then the optimization problem in Eq. \ref{eq:dart_max} becomes 
\begin{align*}
    \hat \epsilon & = \argmin_{\epsilon} - E_{p(\xi | \pi_{\theta^*})} \sum_{t = 0}^{T - 1}\log \left[ \pi_{\theta^*}( \pi_{\hat \theta} (\bx_t) | \bx_t, \epsilon )  \right] \\
    & = \argmin_{\epsilon} - E_{p(\xi | \pi_{\theta^*})} J(\hat \theta, \theta^* | \xi) \log \left[ \frac{\epsilon}{K - 1} \right] + (T - J(\hat \theta, \theta^* | \xi))\log \left[ 1 - \epsilon \right]  \\
\end{align*}
The previous line follows from the fact that along any given trajectory $\xi$ from $\pi_{\theta^*}$, the probability of $\xi$ can be written using only $J(\hat \theta, \theta^* | \xi)$, which is the number times that $\pi_{\hat \theta}$ and $\pi_{\theta^*}$ disagree along $\xi$. Note that this is convex in $\epsilon$. By taking the derivative with respect to $\epsilon$ and setting it equal to zero, we get 
\begin{align*}
    \frac{E_{p(\xi | \pi_{\theta^*})} J(\hat \theta, \theta^* | \xi)}{\epsilon} - \frac{T - E_{p(\xi | \pi_{\theta^*})} J(\hat \theta, \theta^* | \xi)}{1 - \epsilon} & = 0.
\end{align*}
Therefore, we have
\begin{align*}
    \hat \epsilon & = \frac{1}{T}E_{p(\xi | \pi_{\theta^*})} J(\hat \theta, \theta^* | \xi).
\end{align*}

\textbf{Gaussian Scaling}
Given an optimized covaraince matrix $\hat{\Sigma}$, we wish to scale it to some prior over what we expect the robot's final error to be. 

\begin{align*}
&\underset{\beta\geq 0}{\mbox{argmin}} \: |\alpha- E_{p(\xi|\pi_{\theta^*}, \beta*\hat{\Sigma})} \sum^{T-1}_{t=0} l(\bu_t, \pi_{\theta^*}(\bx_t)))|\\
=& \underset{\beta\geq 0}{\mbox{argmin}} \: |\alpha- E_{p(\xi|\pi_{\theta^*}, \beta*\hat{\Sigma})} \sum^{T-1}_{t=0} ||\bu_t- \pi_{\theta^*}(\bx_t)))||^2_2\\
=& \underset{\beta\geq 0}{\mbox{argmin}} \: |\alpha- \beta T\mbox{tr}(\hat{\Sigma})|\\
\end{align*}
 Where the last line followed from known properties of the multi-variate Gaussian and the fact that the level of noise is independent of the current state $\bx_t$. We can derive a solution for $\beta$. 
 
\begin{align*}
=& \frac{\alpha}{T\mbox{tr}(\hat{\Sigma})}
\end{align*}

\subsection{Additional Information for MuJoCo Experiments}
The following hyperparameters were used for all domains. For DAgger and DAgger-B, we swept several values for $\beta$, the stochastic mixing hyperparameter, and found that setting $\beta = 0.5$ yields the best results. For noise injection algorithms, we chose $\alpha= T\mbox{tr}(\hat{\Sigma}_k)$, which corresponds to no additional knowledge being used. For Isotropic-Gaussian noise injection, we set $\Sigma_k^\alpha = I$ for all $k$. Additionally, the covariance matrix for DART was always estimated using held-out demonstrations collected in prior iterations.

In the Walker, Hopper and HalfCheetah domains, we ran these algorithms for 20 iterations, evaluating the learners at 5, 10, 15, and 20 iterations. One initial supervisor demonstration was always collected in the first iteration. To remain consistent, we updated DAgger-B and DART at the same iterations: iteration 2 and iteration 8.

In the Humanoid domain, we ran the algorithms for 200 iterations and evaluated the learners at iterations 50, 100, 150, and 200. Again, we collected one initial supervisor demonstration for DAgger and Isotropic-Gaussian. For DAgger-B and DART, we collected 50 initial supervisor demonstrations before retraining every 25 iterations.

\subsection{Additional Experiments in MuJoCo Domains}
To better understand how DART was able to achieve the reported performance, we asked the additional questions: 1) Does DART reduce the covariate shift? 2) How well do other noise injection methods perform against DART?

\begin{figure}[h]
\includegraphics{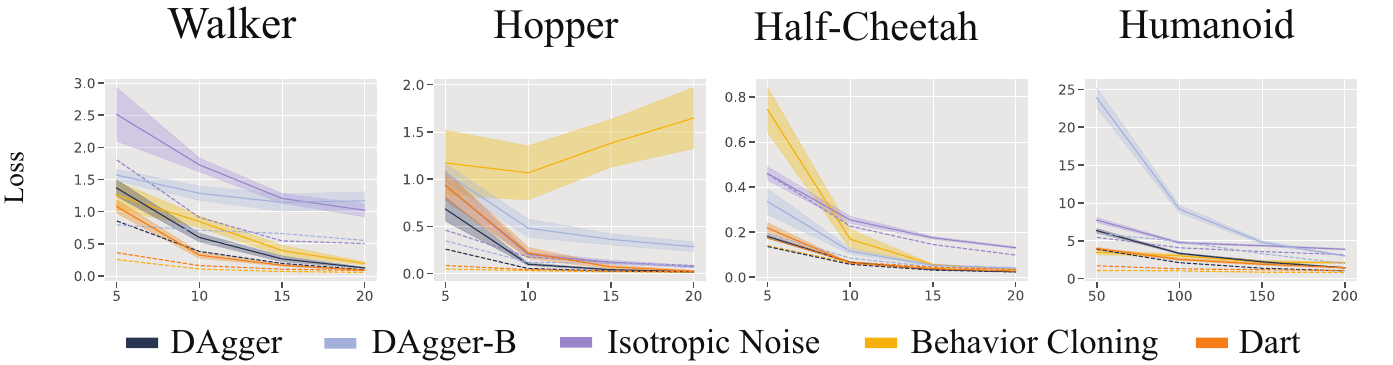}
\centering
\caption{
    \footnotesize
The covariate shift on the four different locomotive domains in MuJoCo we evaluated DART on: Walker, Hopper, Half-Cheetah and Humanoid. The dashed lines correspond to the surrogate losses on the supervisor distributions, $E_{p(\xi |\pi_{\theta^*},\psi^\alpha_k)} J(\theta^R,\theta^*|\xi)$, and the solid lines correspond to the loss on the robot distributions $E_{p(\xi |\pi_{\theta^R})} J(\theta^R,\theta^*|\xi)$. DART is able to reduce the covariate shift in all four domains, which is illustrated by the dash line converging to the solid line. Additionally, DAgger is also able to reduce the shift. However, Behavior Cloning is less effective at doing so, which is illustrated by a slower convergence rate. }

\label{fig:mujocoloss}
\end{figure}

\noindent \textbf{Reducing Covariate Shift} To test how effective DART was at reducing the shift, we evaluate the surrogate loss on both the supervisor's distribution $p(\xi | \pi_{\theta}^*, \hat \psi)$ and the robot's distribution $p(\xi | \pi_{\theta}^*)$. In Fig. \ref{fig:mujocoloss}, we report the losses for each algorithm. Both DAgger and DART are capable of reducing the covariate shift, which is reflected by the smaller disparity between the losses on the supervisor distributions and those on the robot distributions.

\begin{figure}[h]
\includegraphics{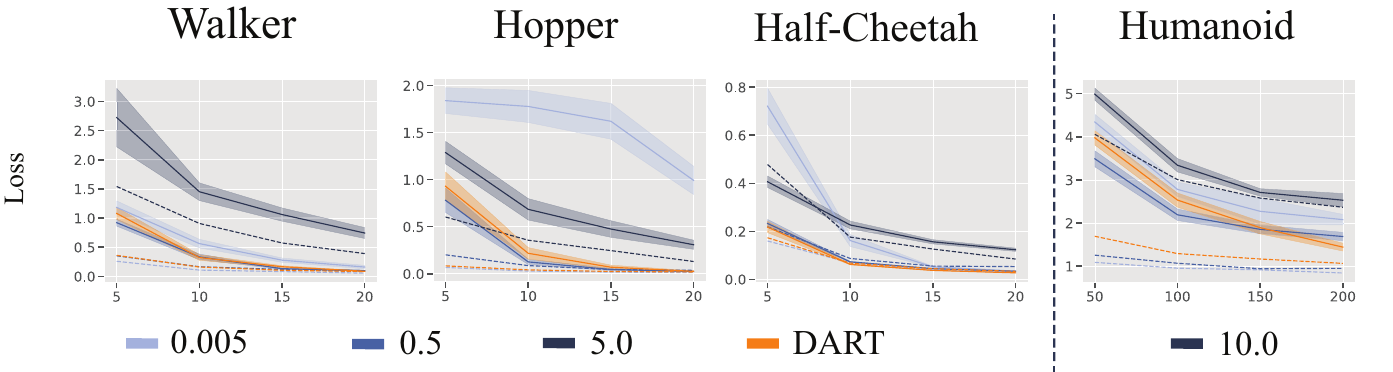}
\centering
\caption{
    \footnotesize
The losses of learners trained from supervisors with randomly sampled covariance matrices are compared against the losses of DART. Again, the dashed lines represent the loss on the supervisor disributions while the solid lines represent the loss on the robot distributions. When the simulated error is well-chose, such as $\text{Tr}(\Sigma) = 0.5$ in these experiments, the performance matches that of DART and the covariate shift is reduced. However higher or lower levels of the noise can cause drastically higher losses.}
\vspace*{-15pt}
\label{fig:mujocorand}
\end{figure}

\noindent \textbf{Random Covariance Matrices} We also compared DART against learners that obtained data from supervisors with random covariance matrices. We define the simulated error of a noisy supervisor to be the expected error of the supervisor on its noisy distribution. In the Gaussian case, note that the expected squared L2 error that the noisy supervisor simulates is equivalent to the trace of the covariance matrix. That is, 
\begin{align*}
    E_{p(\xi | \pi_{\theta^*}, \hat \psi)} \left[ \frac{1}{T} \sum_{t = 0}^{T - 1} \|\bu_t- \pi_{\theta^*}(\bx_t) \|_2^2\right]  & = \text{tr}\ E_{p(\xi | \pi_{\theta^*}, \hat \psi)} \left[ \frac{1}{T} \sum_{t = 0}^{T - 1} (\bu_t- \pi_{\theta^*}(\bx_t))(\bu_t- \pi_{\theta^*}(\bx_t))^T\right] \\
    & = \text{tr}\ \left(\Sigma\right),
\end{align*}
where $\hat \psi$ corresponds to the parameters of the multivariate Gaussian distribution with zero mean and covariance matrix $\Sigma$. Thus, the amount of simulated error can be tuned by tuning the trace of the covariance matrix. We used this fact generate random covariance matrices drawn from the Wishart distribution and scaled the traces of the covariance matrices such that they simulated losses of 0.005, 0.5, and 5.0 for Walker, Hopper, and HalfCheetah and 0.005, 0.5, and 10.0 for Humanoid. The results, reported in Fig. \ref{fig:mujocorand}, suggest that, as long as the simulated error is carefully chosen, randomly sampled covariance matrices may perform as well as DART; however it is not often known in practice what the simulated error should be before training the learner and evaluating it.

\subsection{Additional Information on Robot Setup}
During policy evaluation, we measured success as 1) the robot is able to identify the goal object and 2) there being a clear path between its gripper and the object. Once these conditions are identified an open-loop motion plan is generated to execute a grasp around the target object. In order for the robot to identify the target object, we use the Faster R-CNN trained on the PASCAL VOC 2012 dataset, which has a bottle object category~\cite{NIPS2015_5638}. Once the mustard bottle has been identified and the a path is cleared a human supervisor tells the robot to execute the open-loop motion plan towards a fix position that the goal object is at. The human supervisor is used to ensure that the success criteria is not prone to error in the system, but the supervisor is blinded to which policy is being evaluated to prevent biases. We evaluate the policy 20 times on different sampled initial states.
A video of the learned policy and the robot setup can be found here: \url{https://youtu.be/LfMD69llesg}.



\end{document}